\pdfoutput=1

\documentclass[11pt]{article}

\usepackage[final]{acl}

\usepackage{times}
\usepackage{latexsym}

\usepackage{amsmath} 

\usepackage[T1]{fontenc}

\usepackage[utf8]{inputenc}

\usepackage{microtype}

\usepackage{inconsolata}

\usepackage{graphicx}

\usepackage{graphicx} 
\usepackage{times}
\usepackage{latexsym}
\usepackage[utf8]{inputenc}
\usepackage[T1]{fontenc} 
\usepackage{hyperref}   
\usepackage{url}    
\usepackage{booktabs}   
\usepackage{amsfonts}  
\usepackage{nicefrac}   
\usepackage{microtype}   
\usepackage{lipsum}
\usepackage{fancyhdr} 
\usepackage{graphicx} 
\graphicspath{{media/}}  
\usepackage{makecell}
\usepackage{colortbl}
\usepackage{xcolor}
\usepackage{array}
\usepackage{pifont}
\usepackage{booktabs}
\usepackage{multirow}
\usepackage{hyperref}
\usepackage{url}
\usepackage{wasysym}
\usepackage{tikz}
\usetikzlibrary{shapes}
\usepackage{pgfplots}

\usepackage{amssymb}
\usepackage{enumerate}
\usepackage{wrapfig}
\usepackage[T1]{fontenc}
\usepackage{subfigure}

\usepackage{tcolorbox}

\title{DynaCode: A Dynamic Complexity-Aware Code Benchmark for Evaluating Large Language Models in Code Generation}

\author{
  Wenhao Hu$^1$ ~~~~ Jinhao Duan$^2$  ~~~~ Chunchen Wei$^1$ ~~~~  Li Zhang$^2$ \\
  {\bf ~~~~  Yue Zhang$^2$  ~~~~ Kaidi Xu$^2$\thanks{~~Corresponding author: Kaidi Xu <kx46@drexel.edu>.}} \\
   \\
  $^1$University of Electronic Science and Technology of China\\
  $^2$Drexel University \\
}

\begin{document}
\maketitle
\begin{abstract}
The rapid advancement of large language models (LLMs) has significantly improved their performance in code generation tasks. However, existing code benchmarks remain static, consisting of fixed datasets with predefined problems. This makes them vulnerable to memorization during training, where LLMs recall specific test cases instead of generalizing to new problems, leading to data contamination and unreliable evaluation results.
To address these issues, we introduce DynaCode, a dynamic, complexity-aware benchmark that overcomes the limitations of static datasets. DynaCode evaluates LLMs systematically using a complexity-aware metric, incorporating both code complexity and call-graph structures.
DynaCode achieves large-scale diversity, generating up to 189 million unique nested code problems across four distinct levels of code complexity, referred to as units, and 16 types of call graphs. 
Results on 12 latest LLMs show an average performance drop of $16.8\%$ to $45.7\%$ compared to MBPP+, a static code generation benchmark, with performance progressively decreasing as complexity increases. This demonstrates DynaCode’s ability to effectively differentiate LLMs. Additionally, by leveraging call graphs, we gain insights into LLM behavior, particularly their preference for handling subfunction interactions within nested code.
Our benchmark and evaluation code are available at \href{https://github.com/HWH-2000/DynaCode}{https://github.com/HWH-2000/DynaCode}.
\end{abstract}

\section{Introduction}
The performance of Large Language Models (LLMs) in code generation has garnered significant attention~\cite{10.1145/3695988}. With their powerful language comprehension abilities, LLMs are now capable of autonomously generating high-quality code and, to some extent, addressing complex programming challenges. These advancements have not only accelerated the software development process but have also had a profound impact on enhancing developer productivity~\cite{10.1145/3520312.3534864}. However, as LLMs advance, reliable benchmarks for code generation have become increasingly crucial for evaluating and selecting suitable LLMs.
\begin{figure}[!t]
    \centering
    \includegraphics[width=1.0\columnwidth]{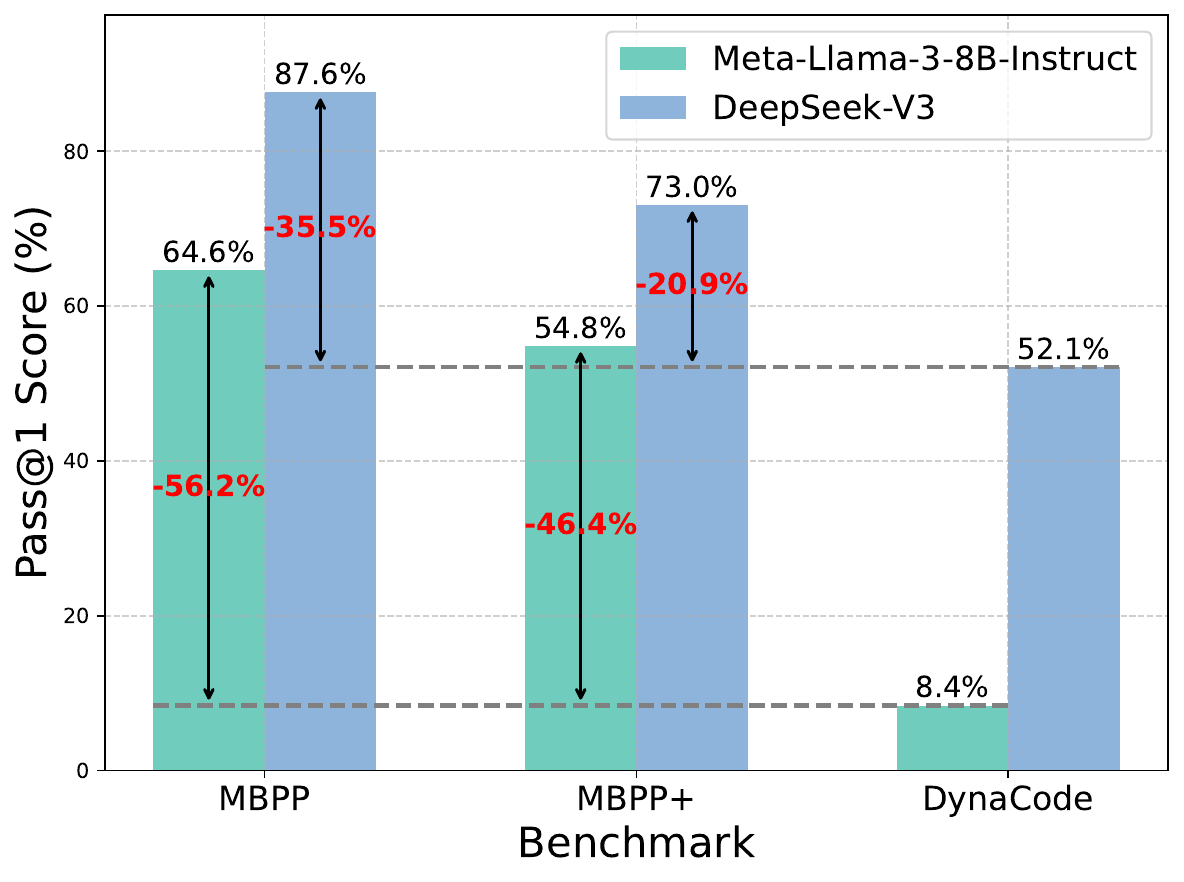}
    \caption{Data contamination on the popular benchmarks MBPP and MBPP+. Meta-Llama-3-8B-Instruct exhibits a significant performance drop from MBPP and MBPP+ to DynaCode.}
    \label{fig:contamination}
    \vspace{-3mm}
\end{figure}
Currently, the evaluation of LLMs' code generation capabilities primarily relies on standardized benchmarks such as HumanEval~\cite{chen2021codex}, MBPP~\cite{austin2021program}, CodeXGLUE~\cite{lu2021codexglue}, and ClassEval~\cite{du2023classeval}. These benchmarks provide an initial reference for assessing the performance of LLMs in code generation by evaluating the functionality and correctness of the generated code. Moreover, recent work such as EvalPlus~\cite{10.5555/3666122.3667065}, 
BigCodeBench~\cite{zhuo2024bigcodebench}, CRUXEval~\cite{gu2024cruxeval}, and EvoEval~\cite{evoeval} aim to enhance evaluation quality by expanding test cases and employing techniques like prompt transformation to convert prompts into more appropriate ones for more precise evaluation.

Despite these developments, existing benchmarks exhibit two notable limitations:

\noindent\textbf{Data Contamination. } 
Existing benchmarks are static and small-scale, making them easily accessible during training and allowing models to memorize test cases instead of generalizing to unseen problems.
Meta-Llama-3-8B-Instruct~\cite{meta_llama3_8b_instruct} and Phi-2~\cite{phi_2} have been reported to exhibit data contamination~\cite{zhang2024careful}, suggesting that the model may ``memorize'' specific test cases or code snippets, which could compromise the accuracy and fairness of the evaluation process.

\noindent\textbf{Uncontrollable Complexity. } 
Existing benchmarks lack systematic complexity control, making it challenging to evaluate LLM performance across different task complexities. While some works~\cite{yu2024codereval, liu2024no} define code complexity using simple metrics such as lines of code and time complexity, these measures fail to capture deep nesting and complex execution dependencies, thereby leaving a critical gap in assessing real-world code generation capabilities.

To address these limitations, we propose DynaCode, a novel dynamic evaluation framework that automatically creates Python code benchmarks by classifying code problems based on complexity and forming nested problems using call graphs. This benchmark provides a more comprehensive and fair evaluation of code generated by LLMs. 
Specifically, DynaCode categorizes code problems into multiple code problem units and, for each unit, constructs call-graph structures of varying complexity. In doing so, it establishes \textit{complexity-aware} metrics along two dimensions: code complexity and call-graph complexity. 
Compared to traditional static benchmarks, DynaCode offers a significantly more diverse and complex evaluation. As shown in Figure~\ref{fig:contamination}, Meta-Llama-3-8B-Instruct~\cite{meta_llama3_8b_instruct}, a model that exhibits data contamination, shows a larger performance drop from MBPP and MBPP+ to DynaCode. DynaCode generating approximately 189 million unique code generation tasks across 4 units of code complexity and 16 call-graph structures. By assessing LLMs from both code complexity and call-graph complexity perspectives, DynaCode provides a structured and scalable evaluation framework while mitigating data contamination.
To further investigate LLM limitations, we analyzed 4279 error examples, categorizing errors into 3 distinct types. Our analysis reveals that LLMs perform well on sequential call graphs but struggle with complex, multi-branch dependencies, highlighting their difficulty in handling deeply nested execution flows and long-range function interactions.
In summary, our major contributions are listed as follows:
\begin{itemize}
    \item We propose a dynamic evaluation strategy that simulates the actual execution process by combining multiple code problems, thereby enabling a fairer and more comprehensive evaluation.
    \item We design complexity-aware metrics combining code complexity and call graphs, integrating static analysis and dynamic execution to create a multidimensional complexity evaluation system with categorized benchmarks.
    \item We introduce DynaCode, a new code generation benchmark, and evaluate multiple LLMs, providing a thorough analysis of its practical utility.
\end{itemize}

\section{Related Works}
\subsection{Dynamic Evaluation}
Recently, growing interest in dynamic evaluation methods has emerged to address data contamination. Several works have focused on different aspects of this challenge. For example, DyVal~\cite{zhu2023dyval} utilizes a graph-based approach to dynamically generate evaluation samples with controllable complexity; NPHardEval~\cite{fan2023nphardeval} generates new evaluation samples for NP-hard mathematical problems; DyVal2~\cite{zhu2024dyval} leverages a probing agent based on LLMs to transform existing problems into new ones, while a judgment agent verifies the generated evaluation samples. Additionally, Benchmark Self-Evolving~\cite{wang2024benchmark} modifies the context or the problem itself, along with its corresponding answers, to reconstruct existing benchmark instances into new variants for dynamic evaluation. DARG~\cite{zhang2024darg} also utilizes LLMs to build reasoning graphs for problems and applies fine-grained graph perturbations across various dimensions. 
However, these methods mainly focus on reasoning domains like logic and mathematics and may not extend well to code generation. Moreover, they rely on LLMs as agents to refine benchmarks, introducing evaluation instability and extra costs.
To address these limitations, this paper proposes a dynamic evaluation strategy tailored for code generation, which automatically creates benchmarks and offers a more detailed and comprehensive assessment.

\subsection{Coding benchmark for LLMs}
The rapid development of LLMs has driven the continuous evolution of code generation benchmarks. Benchmarks such as HumanEval~\cite{chen2021codex} and MBPP~\cite{austin2021program} primarily evaluate LLMs on simple, isolated Python functions, offering a evaluation of code generation capabilities. 
As LLMs have improved, new benchmarks have been continuously proposed, expanding to address higher levels of difficulty~\cite{zhuo2024bigcodebench} and a wider range of programming languages~\cite{zheng2023codegeex, khan2024xcodeeval, cassano2023multipl, ding2023crosscodeeval}. These benchmarks also tackle more complex tasks such as program repair and code reasoning~\cite{liu2024codemind, gu2024cruxeval, jain2024livecodebench}.
In addition, new benchmarks like SWE-Bench~\cite{jimenez2023swe} and EvoCodeBench~\cite{li2024evocodebench} focus on real-world tasks and code evolution, pushing forward the performance evaluation of LLMs in practical applications.
However, existing benchmarks remain static and fixed, lacking systematic complexity control in generated code problems. They typically assess LLMs based on overall performance, failing to provide granular insights into varying code complexities.
We improve this by proposing a complexity-aware metric, allowing for a more precise and systematic evaluation of LLMs' performance.

\section{Methodology}

\begin{figure*}[t]
\centering
\includegraphics[width=0.95\textwidth]{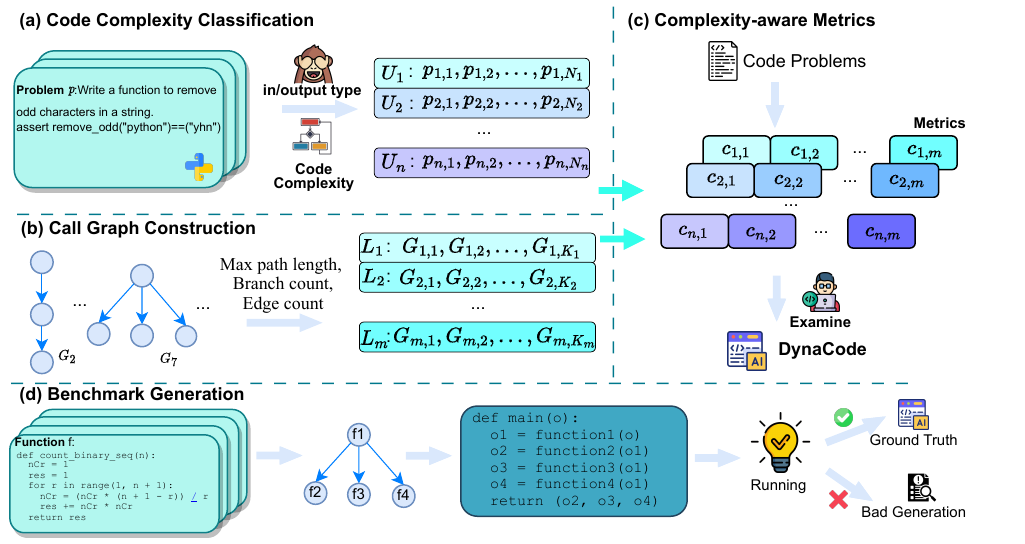}
\vspace{-0.1in}
\caption{
Overview of our proposed DynaCode. (a) Classification of code complexity, resulting in code problems with varying levels of complexity. (b) Construction of function call graphs, categorized based on graph features. (c) Integration of code complexity and graph complexity to form two-dimensional complexity-aware metrics. (d) Benchmark generation process.
}
\label{fig:framework}
\vspace{-0.15in}
\end{figure*}

In this section, we present the construction approach and process of DynaCode, as shown in Figure~\ref{fig:framework}. Specifically, we first introduce the dynamic evaluation strategy based on call graphs in Sec.~\ref{sec:3.1}, explaining how they capture the relationships and dependencies within a program. 
In Sec.~\ref{sec:3.2}, we introduce a complexity matrix that measures code and graph complexity, essential for evaluating tasks of varying difficulty and capturing program interactions.
Finally, we describe the process of generating benchmarks in DynaCode in Sec.~\ref{sec:3.3}.

\subsection{Dynamic Evaluation Strategy
Based on Call Graphs}
\label{sec:3.1}
To comprehensively evaluate the performance of LLMs in code generation tasks, this section introduces a dynamic evaluation strategy based on call graphs. This strategy measures execution performance of generated code from two dimensions: static complexity classification and dynamic code execution behavior. Specifically, this section is divided into two parts: code complexity classification and call graph construction. These parts describe how code problems are classified based on code complexity and how the call graphs are constructed.

\subsubsection{Code Complexity Classification}

Current LLMs generally perform relatively well on code problems involving basic syntax and simple logical structures, but their performance is inconsistent on problems that involve more control flow branches, nested structures, and recursive calls~\cite{jiang2025can, beger2025coconut}. 
To evaluate the code generation capabilities of LLMs more thoroughly, DynaCode first classifies the complexity of existing code problems based on the structural properties of their ground truth code, as shown in Figure~\ref{fig:framework}(a). Methods for classifying code problem complexity include lines of code, cyclomatic complexity~\cite{mccabe1976complexity}, and halstead complexity~\cite{halstead1977elements}. 
Given that LLMs often struggle with code generation tasks involving complex control flow and branching logic, we use cyclomatic complexity to assess code difficulty. It measures the number of independent paths in the control flow, capturing the complexity of branches and loops.
Specifically, for each problem $p_i$ in the set $\mathcal{P}$ of code problems, we use the static code analysis tool Radon~\cite{rubik_radon} to calculate the cyclomatic complexity of the corresponding ground truth code for each problem, denoted as $\nu_{p_i}$, where \( p_i \) represents the index of the problem in the set $\mathcal{P}$. The cyclomatic complexity is calculated using the following formula:
\begin{equation}
    \nu_{p_i} = E_{p_i} - N_{p_i} + 2P_{p_i},
\end{equation}
where \( E_{p_i} \) is the number of edges in the control flow graph, \( N_{p_i} \) is the number of nodes, and \( P_{p_i} \) is the number of connected components. Control flow graphs and cyclomatic complexity computations are shown in Appendix~\ref{app:cyclomatic_complexity_details}.

Based on the calculated cyclomatic complexity values, code problems are classified into different complexity units. 
We define the code complexity as \( U_j \), where \( j \in \{1, 2, \dots, n\} \), representing different cyclomatic complexity units. The set of problems at complexity level \( U_j \) is denoted by:
\begin{equation}
U_j = \{ p_i \mid \alpha_{j-1} \leq \nu_i \leq \alpha_j \}, 
\end{equation}

where:

\begin{enumerate}
    \item \( p_i \) represents a problem with cyclomatic complexity \( \nu_i \),
    \item \( \alpha_{j-1} \) and \( \alpha_j \) are predefined complexity thresholds for the \( j \)-th Unit.
\end{enumerate}

Thus, for each \( j \), \( U_j \) is the set of problems \( p_i \) such that their cyclomatic complexity \( \nu_i \) falls within the interval \( [\alpha_{j-1}, \alpha_j] \), i.e., \( p_i \in U_j \) if and only if \( \alpha_{j-1} \leq \nu_i \leq \alpha_j \).

The choice of thresholds \( \alpha_1, \alpha_2, \dots, \alpha_n \) is based on the distribution of cyclomatic complexity values in code dataset. These thresholds are selected to capture progressively increasing complexity, allowing for a systematic evaluation of LLMs’ performance as task difficulty escalates.

\subsubsection{Call Graph Construction}
A fundamental requirement of execution-based evaluation is to analyze the execution behavior of the generated code~\cite{chen2021codex}.
However, in static benchmarks, where code problems often involve isolated functions, models may memorize specific solutions instead of generalizing, leading to data contamination. To tackle this issue, we use a call-graph structure to construct nested code, as shown in Figure~\ref{fig:framework}(b). After classifying the complexity of code problems, we aim to construct nested code problems and their corresponding nested codes with different structures at the same complexity unit to enhance the diversity of code generation evaluations.
Specifically, we treat each code problem and its ground truth code as nodes in the call graph to form nested problems and nested codes. The call graph is defined as a directed graph \( G = (V, E) \), where the node set \( V \) represents the functions in the code, and the edge set \( E \) represents the call relationships between the functions.
In order to combine code problems into nested problems and corresponding codes, we define the call graph as follows:
\begin{enumerate}
    \item There are exactly \( K \) distinct call-graph structures, where each structure \( G_k = (V_k, E_k) \), for \( k \in \{1, 2, \dots, K\} \), corresponds to a unique function call pattern.
    \item The number of root nodes is \( |V_{\text{root}}| = 1 \), with the root node denoted by \( v_1 \).
    \item The graph is acyclic: for any pair of nodes \( u, v \in V \), if \( u \to v \), then there is no \( v \to u \), i.e., \( G \) is a Directed Acyclic Graph.
\end{enumerate} 
These call-graph structures range from simple linear calls to more complex configurations involving various branches and call relationships, effectively increasing the logical complexity of the code.
Through this diverse set of call-graph structures, DynaCode is able to simulate a variety of real-world scenarios, assessing the ability of LLMs to generate code of varying complexity. 
Detailed illustrations of all call-graph structures are shown in Appendix~\ref{app:call_graph_details}.

To more precisely evaluate the code generation capability of LLMs, DynaCode categorizes the complexity of different call-graph structures to further assess LLM performance within the same unit. 
The complexity of the call graph is influenced by these key features:

\begin{enumerate}
    \item \textbf{Maximum path length} \( L_{\text{max}}(G) \): The longest path from the root node \( r \) to any other node in \( G \).
    \item \textbf{Branch count} \( B(G) \): The total number of branching points in \( G \), indicating how many functions each function calls.
    \item \textbf{Edge count} \( |E| \): The total number of directed edges in the graph, which quantifies the interdependencies between functions.
\end{enumerate}

To measure the complexity of the graph \( G \), we define a comprehensive feature metric \( \mathcal{M} \):

\begin{equation}
    \mathcal{M}(G) = L_{\text{max}}(G) \times B(G) \times |E|.
\end{equation}
The \(\mathcal{M}\) feature, computed as the product of the maximum path length, branch count, and edge count, comprehensively reflects the overall complexity of the call graph. Based on this feature, we categorize the complexity of call graphs into different levels according to predefined thresholds. Let \(\beta_0, \beta_1, \ldots, \beta_m\) be the thresholds. Then, the classification is defined as follows:
\begin{equation}
L_j = \{\, G \mid \beta_{j-1} \leq \mathcal{M}(G) \leq \beta_j \,\}, 
\end{equation}
where \(L_j\) represents the set of call graphs whose \(\mathcal{M}\) values fall within the interval \([\beta_{j-1}, \beta_j]\), corresponding to the \(j\)th level of graph complexity.

\subsection{Complexity-aware metrics}
\label{sec:3.2}
By combining code complexity classification with call-graph complexity classification, we propose a comprehensive complexity measurement matrix, as illustrated in Figure~\ref{fig:framework}(c). This matrix provides a two-dimensional framework for evaluating code complexity, integrating both the internal logical structure of functions and their interrelationships within the call graph. Its goal is to enable a holistic assessment of code complexity by considering two critical aspects: the inherent complexity of function logic and the complexity of its interactions within the overall code structure.
The matrix is represented as:
\begin{equation}
   \mathcal{C} = \{ c_{\xi,\eta} \mid 1 \leq \xi \leq n,\; 1 \leq \eta \leq m \}.
\end{equation}

where \( c_{\xi,\eta} \) represents the complexity value at the intersection of code complexity unit \( \xi \) and call-graph complexity level \( \eta \). Specifically, \( \xi \in \{1, 2, \ldots, n\} \) corresponds to the different units of code complexity, and \( \eta \in \{1, 2, \ldots, m\} \) corresponds to the levels of complexity associated with the call-graph structure. The matrix \( \mathcal{C} \) allows for a detailed, two-dimensional assessment of code complexity by evaluating how the internal logic of functions interacts with the overall structure of the code, which can significantly influence the performance and maintainability of generated code.

To support complexity-aware evaluation, we classify code problems by the cyclomatic complexity of their ground truth code and generate nested variants using diverse call-graph structures with varying call-graph complexity. Both dimensions are independently controllable, allowing future extensions with new code problems or more complex call graphs.

\subsection{Benchmark Generation}
\label{sec:3.3}
Through the steps outlined above, we can construct a systematic code generation benchmark. Our approach improves upon existing code problems by dynamically combining them to generate DynaCode, which incorporates both a complexity framework and randomness into the code problems. The benchmark generation process is illustrated in Figure~\ref{fig:framework}(d).
The specific procedure for generating DynaCode is as follows:

\noindent \textbf{Problem Collection. }
We begin by collecting existing code problems to form our DynaCode unit functions. To mitigate data contamination risks, we actively source recently published code problems from the web and incorporate them into our benchmark. This approach allows us to continuously update the code problem set with new releases, thereby alleviating data contamination.

\noindent \textbf{Code Complexity Classification. }  
For each unit function, we evaluate its complexity using cyclomatic complexity and categorize it accordingly, forming multiple code problem units. Simultaneously, we employ the Monkeytype~\cite{instagram_monkeytype} tool to generate corresponding input and output data for each code problem, discarding any data that cannot be integrated into valid nested codes.

\noindent \textbf{Problem Combination. } 
Then, for each unit of code problems, we merge the problems into new nested problems by aligning their input and output types according to the call-graph structure. Concurrently, we automatically assemble the corresponding code. The automatically generated problem prompt is presented in Appendix~\ref{app:prompt}.
    
\noindent \textbf{Testcase Generation. } 
After obtaining code that conforms to the call-graph structure, we inject the input values from the root node of the call graph into the entire nested code and execute it in batches. If any execution errors occur, the generation is classified as a bad generation. We then filter out these bad generations to retain valid nested code, nested problems, and their corresponding test cases. The valid nested code is presented in Appendix~\ref{app:example}.

\section{Evaluation and Analysis}

\begin{table}[t!]
\small
\centering
\resizebox{\linewidth}{!}{ 
\begin{tabular}{l|cccc}
\toprule
\textbf{Type} & \textbf{Unit 1} & \textbf{Unit 2} & \textbf{Unit 3} & \textbf{Unit 4} \\
\midrule
Base     & 153         & 100      & 76       & 76       \\
\midrule
Level 1  & 88,339      & 10,553   & 7,931    & 23,996   \\
Level 2  & 3,300,172   & 157,950  & 107,578  & 585,379  \\
Level 3  & 27,771,290  & 518,215  & 332,610  & 3,428,967 \\
Level 4  & 133,358,131 & 2,528,807 & 1,928,288 & 15,114,935 \\
\bottomrule
\end{tabular}
}
\caption{Number of problems in different units and the total number of generated nested code problems formed by combining them with call-graph structures. Units represent sets of code problems categorized by cyclomatic complexity, while Levels indicate the structural complexity of call graphs.}
\label{statistics}
\vspace{-6mm}
\end{table}

\subsection{Experimental Setup}

\begin{table*}[th!]
\definecolor{deepgreen}{RGB}{0,100,0}
  \centering
  
    \small
  \renewcommand{\arraystretch}{0.95}
  \resizebox{\linewidth}{!}{ 
  \begin{tabular}{lr|cc|ccccc}
    \toprule
    
    \multirow{2}{*}{\textbf{Model}} & \multirow{2}{*}{\textbf{Params}}  & \multirow{2}{*}{\textbf{MBPP}} & \multirow{2}{*}{\textbf{MBPP+}}   & \multicolumn{5}{c}{\textbf{DynaCode}} \\
    & & & & \textbf{Unit 1} & \textbf{Unit 2} & \textbf{Unit 3} & \textbf{Unit 4} & \textbf{Average} \\
    \midrule

    GPT-4o & - & 87.6 & 72.2  & $ 74.4~(\pm 1.6)$ & $ 48.7~(\pm 1.4)$ & $ 56.2~(\pm 1.4)$ & $ 42.3~(\pm 0.3)$ & $ 55.4~(\pm 0.9)$ \\
    
    GPT-3.5-Turbo & -  & 82.5 & 69.7 &  $ 34.9~(\pm 1.0)$ & $ 30.5~(\pm 1.6)$ & $ 25.6~(\pm 0.6)$ & $ 26.1~(\pm 1.2)$ & $ 29.3~(\pm 0.4)$ \\
    \midrule 
    
    DeepSeek-V3 & 236B & 87.6  & 73.0 & $ 65.9~(\pm 0.8)$ & $ 41.6~(\pm 2.3)$ & $ 53.6~(\pm 0.7)$ & $ 47.3~(\pm 1.5)$ & $ 52.1~(\pm 0.4)$ \\
    \midrule
    
     Qwen2.5-Coder-32B-Instruct  & 32B& 90.5 & 77.0  &$ 59.3~(\pm 1.6)$ & $ 33.6~(\pm 1.6)$ & $ 44.1~(\pm 1.1)$ & $ 36.0~(\pm 0.8)$ & $ 43.2~(\pm 0.2)$ \\
    \midrule
    WizardLM-2-8x22B & 176B & 71.7 & 60.8 & $ 35.4~(\pm 0.8)$ & $ 27.5~(\pm 0.9)$ & $ 25.1~(\pm 0.8)$ & $ 12.8~(\pm 1.1)$ & $ 25.2~(\pm 0.6)$ \\  \midrule
    
    Mixtral-8x22B-Instruct-v0.1 & 176B & 73.8 & 64.3 & $ 35.4~(\pm 0.8)$ & $ 27.2~(\pm 1.3)$ & $ 25.0~(\pm 0.5)$ & $ 12.8~(\pm 0.5)$ & $ 25.1~(\pm 0.7)$ \\
    \midrule
    
    Phind-CodeLlama-34B-v2 & 34B & 85.4 & 69.6 & $ 40.9~(\pm 0.9)$ & $ 29.5~(\pm 1.8)$ & $ 45.2~(\pm 1.5)$ & $ 25.4~(\pm 0.2)$ & $ 35.3~(\pm 0.6)$ \\
    \midrule
    
     starcoder2-15b-instruct-v0.1 & 15B & 78.0  & 65.1  & $ 40.7~(\pm 1.0)$ & $ 29.4~(\pm 1.4)$ & $ 45.0~(\pm 1.3)$ & $ 25.3~(\pm 0.5)$ & $ 35.1~(\pm 0.4)$ \\
    \midrule

    codegemma-7b-it & 7B & 70.4 & 56.9 & $ 4.6~(\pm 0.4)$ & $ 2.7~(\pm 0.2)$ & $ 1.1~(\pm 0.2)$ & $ 3.0~(\pm 0.4)$ & $ 2.9~(\pm 0.2)$ \\
    \midrule

   Meta-Llama-3.1-405B-Instruct & 405B & 88.4 & 73.0 & $ 49.7~(\pm 0.9)$ & $ 40.0~(\pm 1.6)$ & $ 47.6~(\pm 1.1)$ & $ 26.9~(\pm 1.2)$ & $ 41.0~(\pm 0.8)$ \\
        
     Meta-Llama-3.3-70B-Instruct& 70B & 89.2 & 75.1 & $ 36.0~(\pm 1.4)$ & $ 27.5~(\pm 1.4)$ & $ 54.9~(\pm 1.1)$ & $ 31.2~(\pm 0.8)$ & $ 37.4~(\pm 0.7)$ \\
     
      Meta-Llama-3.1-8B-Instruct& 8B & 68.3 & 55.6 & $ 14.1~(\pm 1.0)$ & $ 9.7~(\pm 0.6)$ & $ 8.4~(\pm 1.0)$ & $ 7.4~(\pm 0.7)$ & $ 9.9~(\pm 0.8)$ \\

    \bottomrule
  \end{tabular}
  }
  \caption{Pass@1 results on MBPP, MBPP+, and our DynaCode across varying code complexities. For DynaCode, results are reported for different function complexity levels. To ensure robustness, all experiments were conducted three times with 5 different random seeds, and the average results are presented.}
  \label{tab1}
\end{table*}

\textbf{Data. } We have selected the MBPP+, processed by EvalPlus~\cite{10.5555/3666122.3667065}, as our unit function set. MBPP+ is an extended version of the MBPP~\cite{austin2021program}, enhanced and refined to provide more comprehensive test cases and solutions. To address the potential of data contamination on unit functions, we also curated a set of the latest programming problems from LeetCode\footnote{LeetCode, \url{https://leetcode.com/}}. These problems, along with their corresponding solutions collected using official test cases, were integrated into our evaluation. As a demonstration, we introduced 22 new code problems into Unit 3 and 18 new code problems into Unit 4. The combination of these two sources forms our unit function set, ensuring both diversity and relevance to real-world scenarios.

\noindent \textbf{Evaluation metric.} Following previous work~\cite{chen2021codex}, we use Pass@1 as the evaluation metric, which measures the percentage of problems solved correctly on the first attempt without further corrections. Since our benchmark introduces progressively increasing complexity levels, we use Pass@1 as a unified evaluation metric to ensure consistent comparisons with MBPP, MBPP+, and across different complexity levels.

\noindent \textbf{Evaluated LLMs. } We selected a range of mainstream LLMs to evaluate the effectiveness of DynaCode in code generation tasks. The evaluation models include GPT-4o~\cite{achiam2023gpt}, GPT-3.5-turbo~\cite{openai_gpt35}, DeepSeek-V3~\cite{liu2024deepseek}, Qwen2.5-Coder-32B-Instruct~\cite{hui2024qwen2}, WizardLM-2-8x22B~\cite{xu2023wizardlm}, Mixtral-8x22B-Instruct-v0.1~\cite{jiang2024mixtral}, Phind-CodeLlama-34B-v2~\cite{roziere2023code}, starcoder2-15b-instruct-v0.1~\cite{wei2024selfcodealign}, codegemma-7b-it~\cite{team2024codegemma}, Meta-Llama-3.1-405B-Instruct~\cite{meta_llama_3.1_405B}, Meta-Llama-3.3-70B-Instruct~\cite{meta_llama_3.3_70B}, and Meta-Llama-3.1-8B-Instruct~\cite{meta_llama_3.1_8B}. To ensure the stability and consistency of the evaluation results, we set the temperature to 0 to eliminate any randomness introduced during the generation process. Our prompts are shown is Appendix~\ref{app:prompt}.

\noindent \textbf{Benchmark Statistics. } We have compiled statistics on the number of problems that can be generated from the selected dataset after applying the dynamic evaluation strategy, as shown in Table~\ref{statistics}. 
Code problems are first partitioned into Units using cyclomatic complexity computed from the ground truth code, with each Unit representing a set of problems of similar complexity. Within each Unit, we construct nested problems by combining these base problems with 16 distinct call-graph structures. To further classify call-graph complexity, the call graphs are grouped into 4 Levels based on features such as branch count. 
Table~\ref{tab:graph_number_details} provides further details on the number of problems generated for each call graph, facilitating an understanding of the scale and distribution of problems across different structures.

\begin{figure*}
    \centering
    \includegraphics[width=0.99\textwidth]{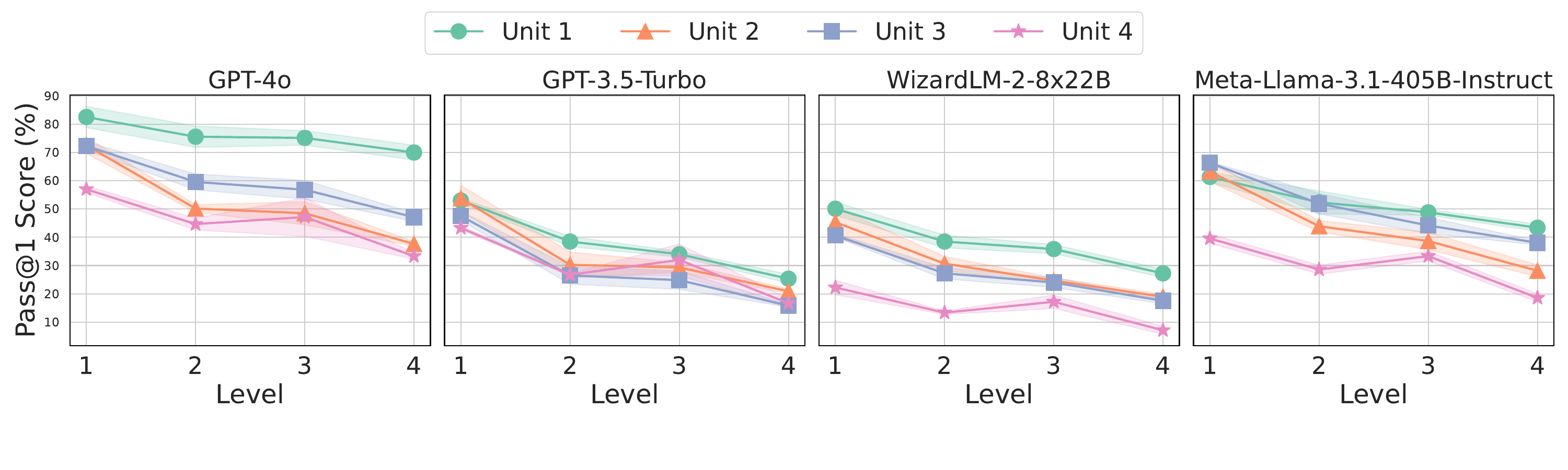}
    \vspace{-6mm}
    \caption{Comparison of Average Pass@1 scores across 4 LLMs (GPT-4o, GPT-3.5-Turbo, WizardLM-2-8x22B, Meta-Llama-3.1-405B-Instruct) at different complexity levels. 
    }
    \label{fig:performance}
    
\end{figure*}

\subsection{Benchmark Results}

\noindent
\textbf{Model Performance.}
The results of our benchmark are summarized in Table~\ref{tab1}. Compared to traditional benchmarks like MBPP and MBPP+, DynaCode shows a more pronounced decline in model performance as code complexity increases.
For example, GPT-4o achieves a Pass@1 score of $87.6\%$ on MBPP and $72.2\%$ on MBPP+, but drops significantly to $55.4\%$ on DynaCode. A similar trend is observed with GPT-3.5-Turbo, which drops from $82.5\%$ on MBPP to $29.3\%$ on DynaCode.
These results highlight the increasing complexity in our benchmark, confirming its ability to assess models on more complex, real-world code generation tasks. Specific performance variations are shown for selected models in Figure~\ref{fig:performance}.
Our complexity-aware metrics effectively differentiate model capabilities, as evidenced by consistent performance degradation across complexity levels. Models like GPT-4o show better resilience to increasing complexity, while others like WizardLM-2-8x22B struggle as complexity rises. This demonstrates DynaCode's robustness in evaluating not just basic code generation but also handling nested, multi-function code structures. The performance trends across all models and complexity scenarios further validate our benchmark’s contribution in revealing the challenges faced by LLMs in complex code generation tasks. We provide detailed performance trends and Pass@3 results for all models, as shown in Appendix~\ref{app:performance_overall}.

\begin{figure}[!t]
    \centering
    \includegraphics[width=0.94\linewidth]{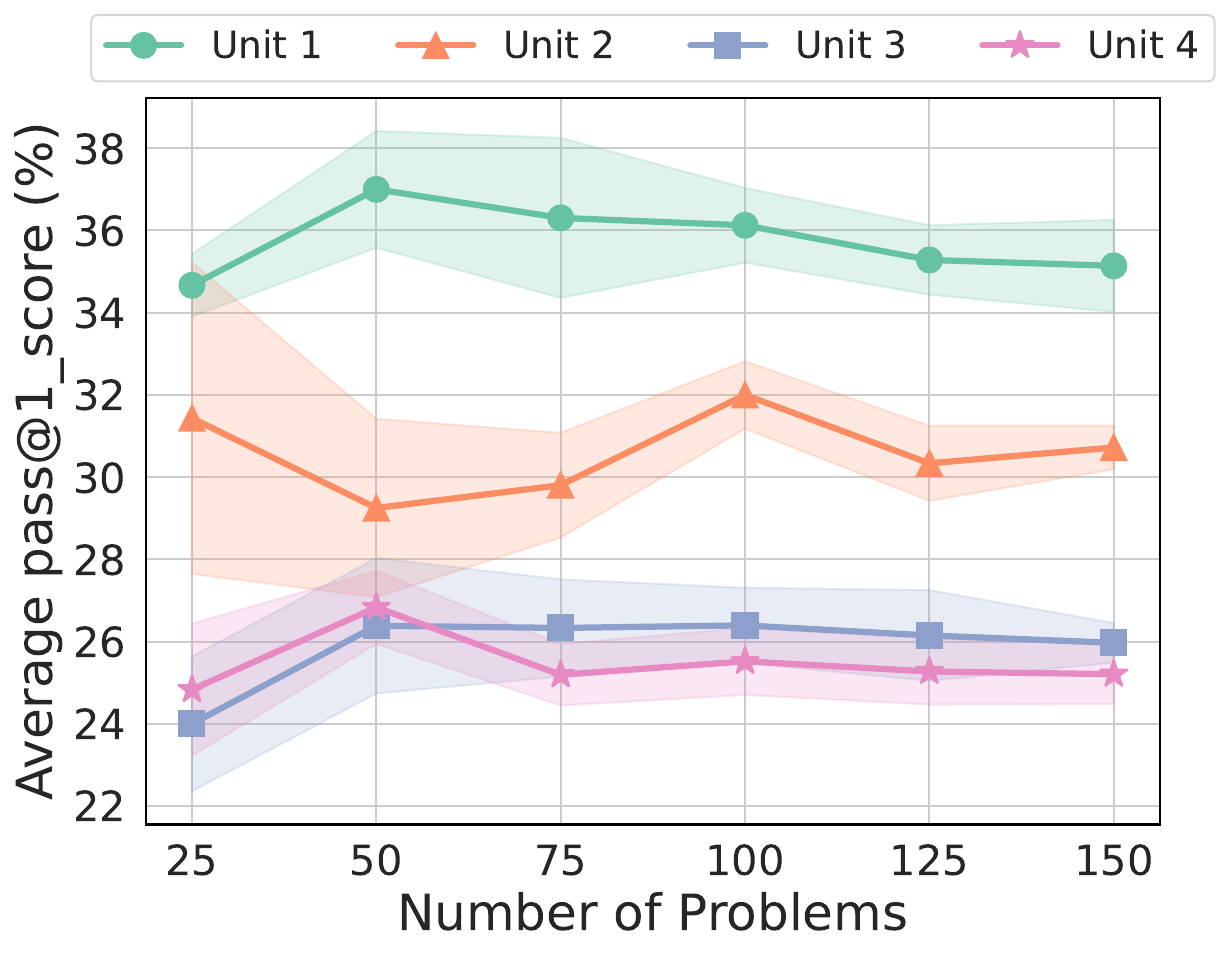}
    \caption{Experiment on the number of problems. For each graph in every unit, a corresponding number of problems is generated.}
    \label{fig:Number}
\end{figure}

\noindent
\textbf{Effect of the Problem Sizes.} 
To further investigate our benchmark, we conducted a study on the effect of the number of dynamically generated problems on the evaluation performance of the benchmark. For each unit and each call graph type, we generated different numbers of problems $\{25, 50, 75, 100, 125, 150\}$, and the results are shown in Figure~\ref{fig:Number}. The experiments indicate that when the number of problems is greater than or equal to $75$, the evaluation results meet our requirements for stability and reliability. Therefore, we set $100$ for other experiments. Despite the variations in the number of generated problems, DynaCode can systematically evaluate code generation tasks at various complexity levels and provide reliable assessments for each complexity scenario. Detailed results for each number are provided in Appendix~\ref{app:performance_number}.

\subsection{Experimental Insights}

\begin{figure}[!t]
    \centering
    \includegraphics[width=0.96\linewidth]{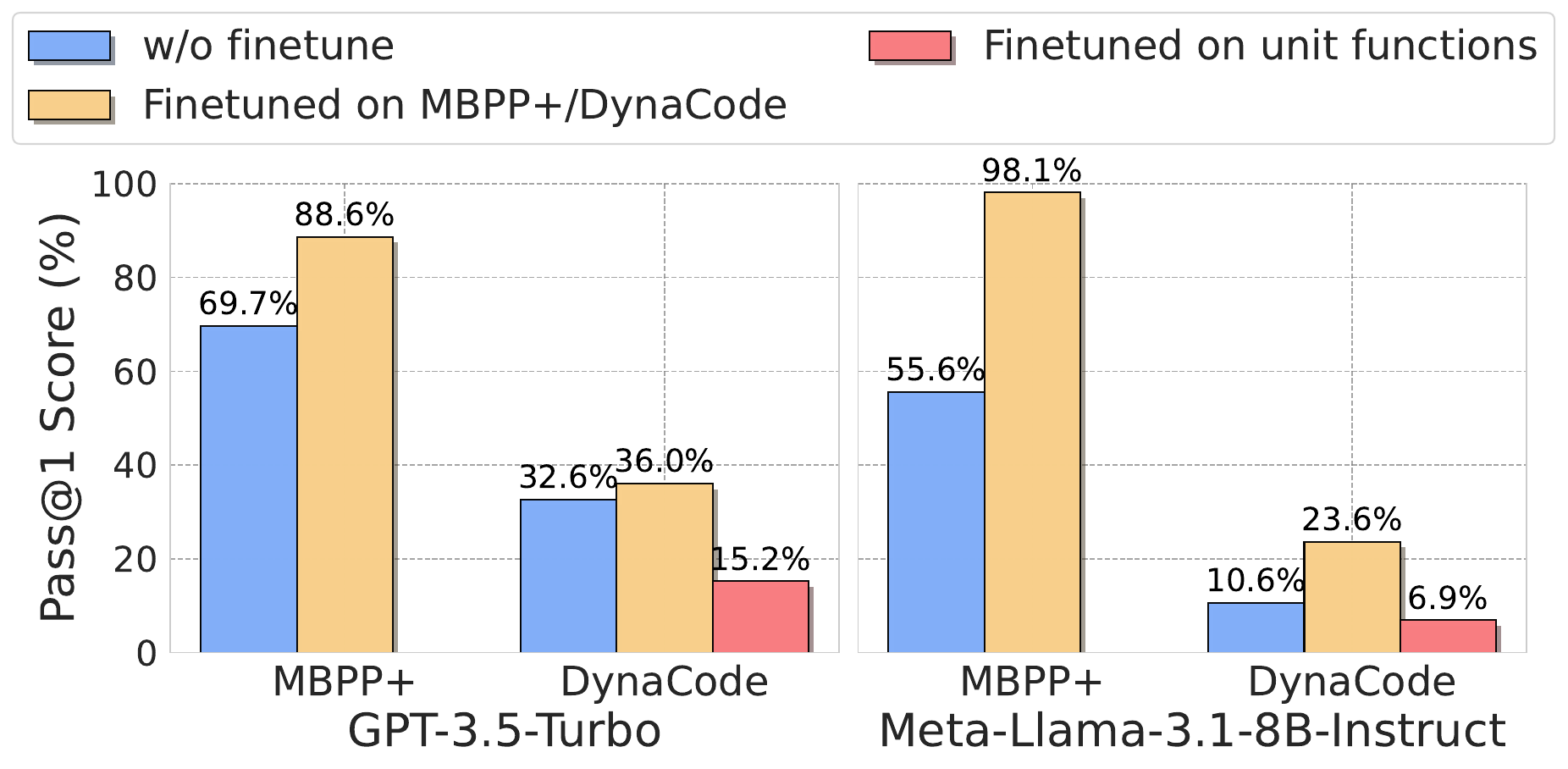}
    \caption{
    Pass@1 scores of GPT-3.5-Turbo and Meta-Llama-3.1-8B-Instruct before and after fine-tuning on MBPP+ and DynaCode, and DynaCode unit functions.
    }
    \label{fig:fineture}
\end{figure}

\begin{figure*}
    \centering
    \includegraphics[width=0.99\textwidth]{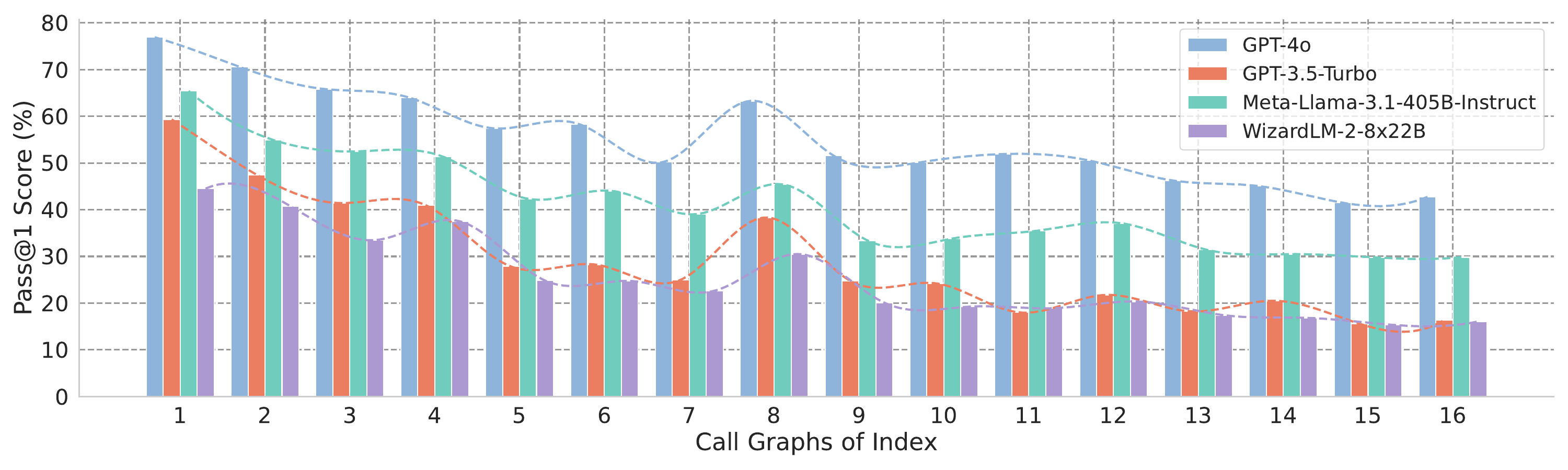}
    \caption{
    Comparative performance evaluation of 4 LLMs (GPT-4o, GPT-3.5-Turbo, Meta-Llama-3.1-405B-Instruct, and WizardLM-2-8x22B) across different call graphs. Models exhibit better performance on sequential call graphs $\{G_1, G_2, G_3, G_4, G_8\}$. Details of the call graphs are provided in Appendix~\ref{app:call_graph_details}.
    }
    \label{fig:graph_performance}
\end{figure*}

\begin{table*}[th!]
\definecolor{deepgreen}{RGB}{0,100,0}
  \centering
    \small
  \renewcommand{\arraystretch}{0.95}
  \resizebox{\linewidth}{!}{ 
  \begin{tabular}{l|l|cccc}
    \toprule
    
    \multirow{2}{*}{\textbf{Ability}}   &  \multirow{2}{*}{\textbf{Error Type}}   & \multicolumn{4}{|c}{\textbf{DynaCode}} \\
     & & \textbf{Unit 1} & \textbf{Unit 2} & \textbf{Unit 3} & \textbf{Unit 4}  \\
    \midrule
    
    Problem Understanding & AssertionError, ValueError, RecursionError, ZeroDivisionError & 64.1\% (615) & 79.9\% (832) & 88.2\% (1027) & 88.8\% (988) \\ 
    \midrule
    
    Code Pattern Generation & SyntaxError, IndentationError  & 6.6\% (63) & 0.4\% (4) & 0.2\% (2) & 1.5\% (17) \\
    \midrule 
    
    Context Management & NameError, AttributeError, TypeError, IndexError, UnboundLocalError & 29.4\% (282) & 19.7\% (205) & 11.7\% (136) & 9.4\% (105) \\
    
    \midrule
    Other &RuntimeError, OverflowError & - & - & - & 0.3\% (3) \\

    \bottomrule
  \end{tabular}
  }
    \caption{Error analysis of GPT-3.5-Turbo on DynaCode. The model was evaluated across 4 units of increasing complexity, with 100 questions per graph. The table summarizes the distribution and frequency of errors categorized by problem understanding, code pattern generation, and context management. 
  }
  \label{tab:error}
\end{table*}

\noindent \textbf{DynaCode Limits Memorization for More Reliable Evaluation.}  
A key challenge in evaluating LLMs is data contamination, where models may memorize training data instead of generalizing effectively. 
To explore this issue, we fine-tuned a commercial and an open-source LLM, GPT-3.5-Turbo, and Meta-Llama-3.1-8B-Instruct, on both the MBPP+ and DynaCode datasets, and then evaluated their performance as shown in Figure~\ref{fig:fineture}.
To ensure a fair comparison and avoid catastrophic forgetting, we propose a fine-tuning strategy as follows: for GPT-3.5-Turbo, we trained on MBPP+ for 5 epochs, on DynaCode unit functions for 5 epochs, and on DynaCode for 1 epoch, while keeping the total fine-tuning steps fixed at 1890; for Meta-Llama-3.1-8B-Instruct, we trained on MBPP+ for 10 epochs, on DynaCode unit functions for 10 epochs, and the DynaCode for 2 epochs, while keeping the total fine-tuning steps fixed at 3780.
The results show a significant improvement in GPT-3.5-Turbo’s Pass@1 score on MBPP+, which increased from 69.7\% to 88.6\%, while on DynaCode, the improvement was much smaller, rising from 32.6\% to 36.0\%. A similar pattern was observed for Meta-Llama-3.1-8B-Instruct, where the model’s performance on MBPP+ surged from 55.6\% to 98.1\%, but only improved slightly from 10.6\% to 23.6\% on DynaCode.
To further investigate the impact of data contamination, we introduced a fine-tuning setting using only the unit functions from DynaCode, and then evaluated the models on the full DynaCode benchmark. Under this setting, GPT-3.5-Turbo’s Pass@1 score dropped to 15.2\%, while Meta-Llama-3.1-8B-Instruct’s score fell to 6.9\%. These smaller gains suggest that the models are not simply memorizing the data, implying that DynaCode’s dynamic evaluation strategy effectively mitigates data contamination. As a result, DynaCode offers a more reliable and comprehensive assessment of LLMs' true code generation capabilities, ensuring that the evaluation is based on the model's generalization ability rather than its capacity to memorize specific training examples.

\noindent \textbf{LLMs Are Good at Sequential Execution.} 
We evaluated 4 LLMs on different call graphs, averaging their Pass@1 scores across all 4 units, as shown in Figure~\ref{fig:graph_performance}. The results reveal a clear trend: LLMs perform significantly better on sequentially structured graphs $\{G_1, G_2, G_3, G_4, G_8\}$, which involve linear execution with minimal branching. This suggests that LLMs excel at straightforward function compositions and stepwise computations. However, as call graph complexity increases, particularly in multi-layered, multi-branch structures $\{G_9, G_{10}, G_{11}, G_{12}, G_{13}, G_{14}, G_{15}, G_{16}\}$, model performance drops considerably. This indicates that LLMs struggle with parallel function dependencies and managing execution across interdependent subfunctions. Among the evaluated models, GPT-4o consistently outperforms others across all graphs, showing greater resilience to structural complexity. Nevertheless, even GPT-4o exhibits a notable decline in high-complexity graphs, highlighting a fundamental limitation in LLMs' ability to generate and manage deeply nested, interdependent code structures.

\noindent \textbf{LLMs Struggle with Problem Understanding as Complexity Increases.} 
We evaluate the code generation capabilities of GPT-3.5-Turbo on DynaCode by analyzing the distribution and frequency of errors across 4 units, classifying them into Problem Understanding, Code Pattern Generation, and Context Management abilities based on common code errors. The results are summarized in Table~\ref{tab:error}. The findings show that the error rate for Problem Understanding increases progressively from $64.1\%$ in unit 1 to $88.8\%$ in unit 4, indicating that the increasing complexity of the code problems makes it more difficult for GPT-3.5-Turbo to correctly understand the problem requirements. In contrast, the error rates for Code Pattern Generation and Context Management show a decreasing trend across units. However, this reduction does not imply an improvement in capabilities. Instead, it reflects a shift in the error distribution: as the task complexity increases, the model often fails at the Problem Understanding stage, preventing it from generating correct code that would expose errors in syntax or context management. The error classification details are shown in Appendix~\ref{app:error}.

\section{Conclusion}

We present DynaCode, a dynamic, complexity-aware benchmark designed to systematically evaluate LLMs on code generation tasks. By integrating code complexity with call-graph structures, DynaCode generates nested code problems that capture varying levels of code complexity and inter-function dependencies. We evaluated 12 of the latest LLMs, and the results reveal significant performance drops as both unit and graph complexity increase, highlighting DynaCode’s ability to systematically assess LLMs. Notably, DynaCode also addresses the challenge of data contamination and demonstrates its capacity to mitigate this limitation. In summary, DynaCode provides a new perspective for examining and analyzing LLMs, offering a scalable framework for more comprehensive and reliable LLM evaluation.

\section*{Limitations}

DynaCode primarily focuses on relatively call-graph structures, with a maximum node count of 5. While this ensures manageable complexity for current LLMs, it is possible that more advanced LLMs in the future may learn to handle such call patterns. We will extend DynaCode to include more complex call-graph structures in future work, further challenging LLMs and enhancing the benchmark’s scalability.

\bibliography{custom}

\appendix

\clearpage
\clearpage
\newpage

\appendix
\section*{Appendix}

\section{Cyclomatic Complexity Details}
\label{app:cyclomatic_complexity_details}

\begin{figure}[!th]
    \centering
    \includegraphics[width=1\linewidth]{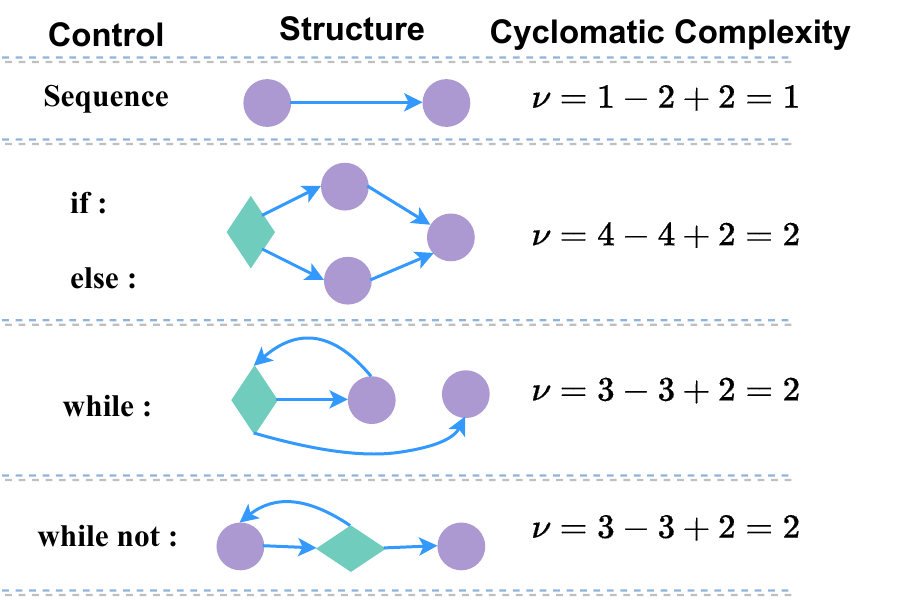}
    \vspace{-3mm}
    \caption{Cyclomatic complexity calculations for Sequence, If-Else, While, and While-Not control structures, illustrating their impact on code complexity.}
    \label{fig:cc}
\end{figure}

We present the cyclomatic complexity calculations for \texttt{Sequence}, \texttt{If-Else}, \texttt{While}, and \texttt{While-Not} control structures, illustrating their impact on code complexity in Figure~\ref{fig:cc}. Cyclomatic complexity quantifies the number of independent execution paths in a program, making it a useful metric for assessing the logical complexity of different control flows. 
Simple structures like \texttt{Sequence} have a lower complexity, as they follow a single execution path, whereas \texttt{If-Else} and \texttt{While} introduce branching and loops, increasing the number of possible execution paths. \texttt{While-Not} adds additional conditional constraints, further influencing complexity. Understanding these calculations helps in evaluating how LLMs handle different levels of logical complexity in generated code.

\begin{figure}[!th]
    \centering
    \includegraphics[width=1\linewidth]{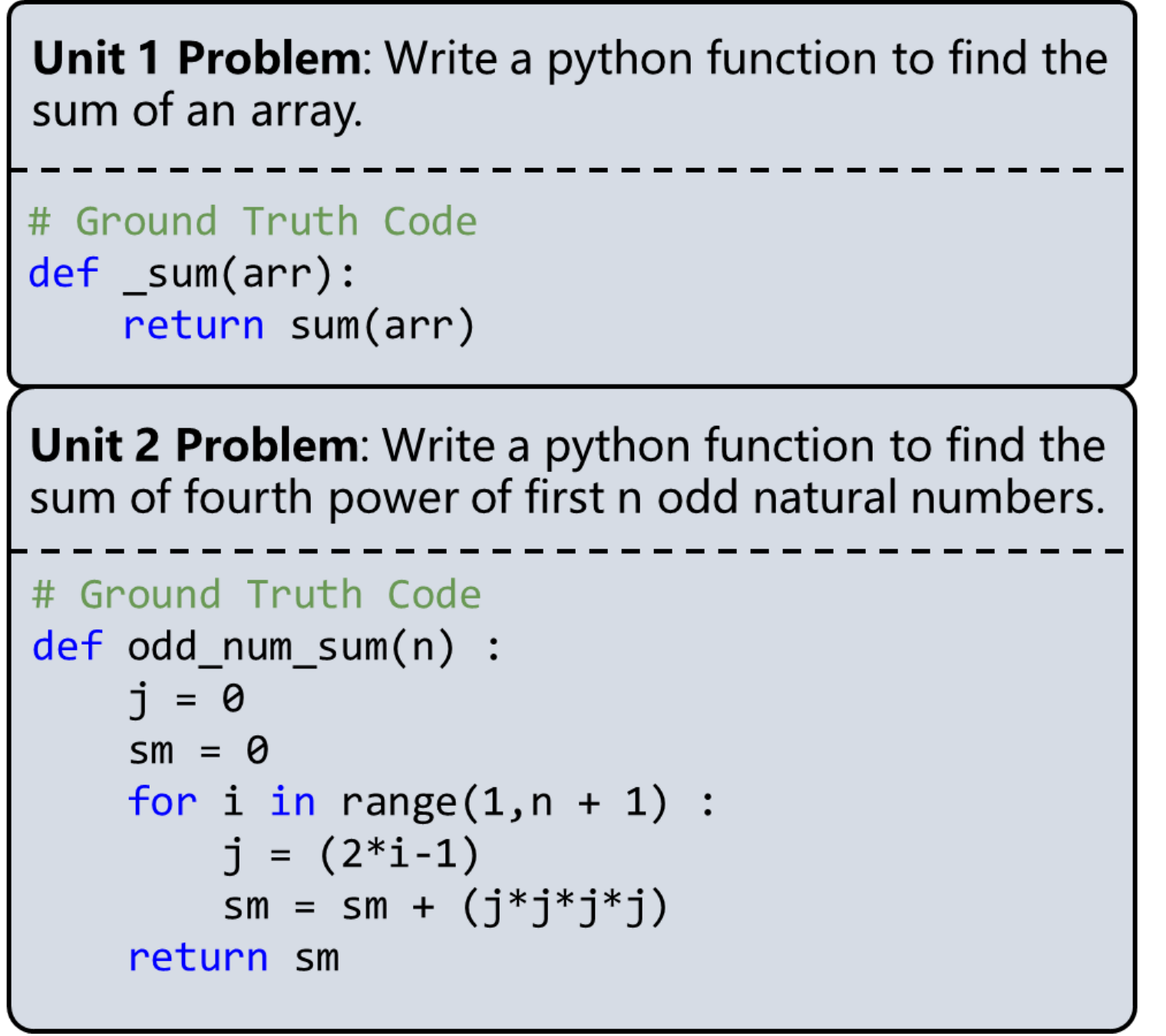}
    \vspace{-3mm}
    \caption{Examples of code problems with increasing complexity. As problem complexity increases from Unit 1 to Unit 2, the corresponding solution also becomes more complex, while maintaining clear and efficient code structure.}
    \label{fig:code problem}
\end{figure}

We further leverage cyclomatic complexity to categorize the difficulty of code problems in our benchmark by computing this metric on the ground truth code. As shown in Figure~\ref{fig:code problem}, the coding problems consistently become more challenging as the complexity increases, enabling a principled evaluation of how LLMs handle progressively harder logical structures.

\section{Call Graph Details}
\label{app:call_graph_details}
\begin{figure*}[t]
\centering
\includegraphics[width=0.95\textwidth]{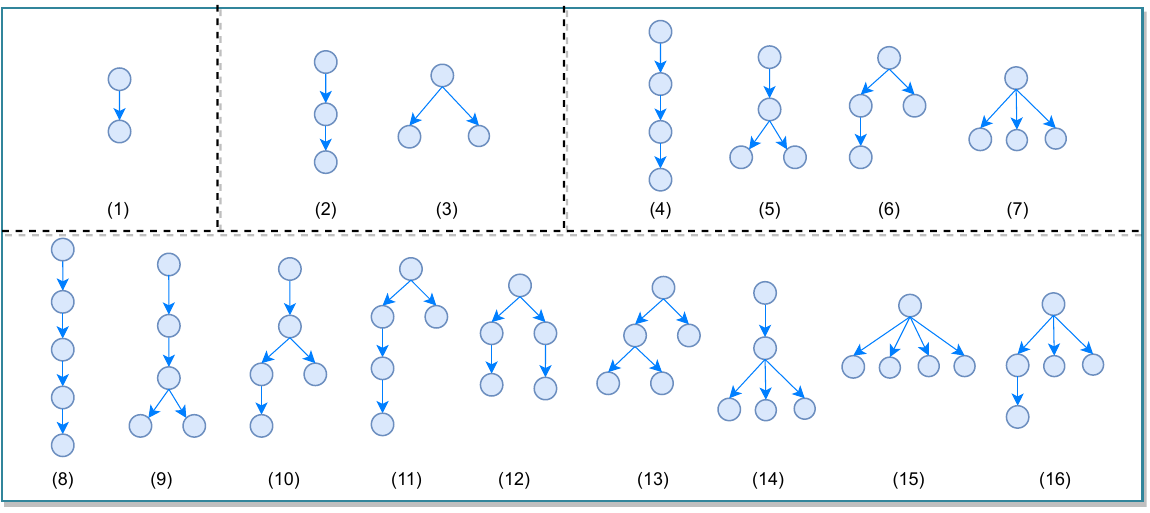}
\caption{
Illustration of 16 distinct call-graph structures used in our experiments, each modeled as a directed acyclic graph with up to 5 nodes and a single root.
}
\label{fig:call_graph}
\end{figure*}

In our experiments, we set the maximum number of nodes in the call graph to 5. This configuration enables the generation of 16 distinct call-graph structures, each corresponding to a unique function call pattern. Every call graph is modeled as a directed acyclic graph with a single root node, ensuring a unified entry point for function calls. Figure~\ref{fig:call_graph} illustrates the detailed configurations of all 16 call-graph structures. These structures provide a comprehensive testbed for evaluating the performance of LLMs in generating code with complex nested function calls, thereby enhancing our assessment of their ability to handle varying levels of logical and structural complexity.

\section{Benchmark Statistics Details}
We report the details of benchmark statistics in Table~\ref{tab:graph_number_details}, which shows the number of problems contained in different units and the corresponding number of benchmarks that can be generated through combinations.
We also compare the total number of problems in DynaCode with those in other code generation benchmarks, as illustrated in the Table~\ref{tab:benchamrk_size}.

\section{Model Performance Details}
\label{app:performance}

\subsection{Overall Results}
\label{app:performance_overall}

\begin{figure*}[t]
\centering
\includegraphics[width=1\textwidth]{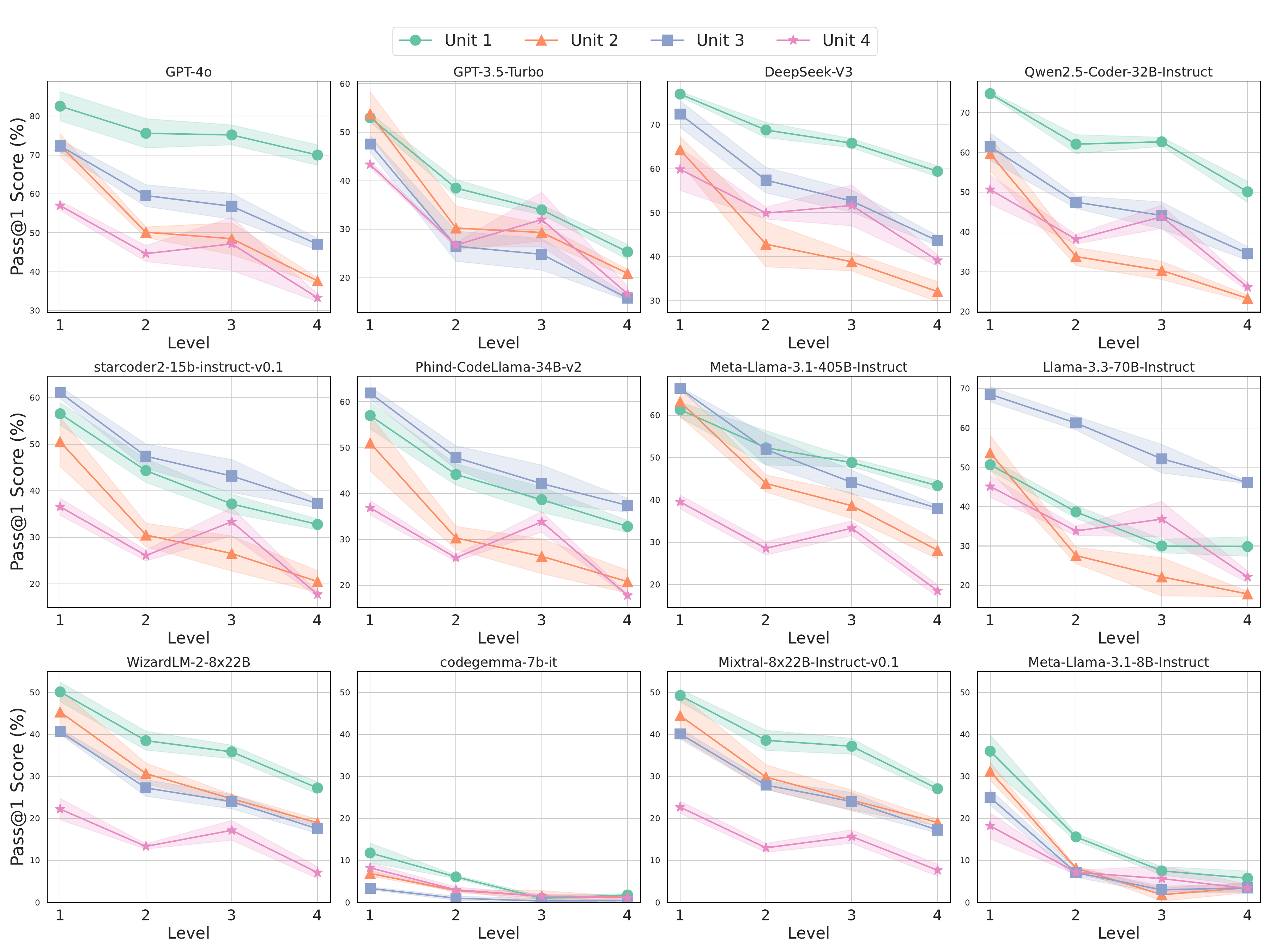}
\caption{
Performance details of various models on DynaCode across different complexity levels. Each subfigure illustrates the Pass@1 score trends for units 1 to 4, highlighting how model performance degrades as code complexity increases.
}
\label{fig:all_performance}
\end{figure*}

Figure~\ref{fig:all_performance} presents the Pass@1 performance details of all evaluated models on DynaCode across varying levels of code complexity. The evaluation reveals consistent performance degradation across all models as code complexity increases, underscoring the benchmark's ability to challenge models beyond basic code generation.

Models like GPT-4o and DeepSeek-V3 exhibit relatively robust performance, maintaining higher Pass@1 scores even at higher complexity levels. In contrast, models such as WizardLM-2-8x22B and codegemma-7b-it show a steep decline in performance, struggling significantly with more complex, nested code structures. These trends align with the observations discussed in the main benchmark results, further validating the robustness of DynaCode in differentiating model capabilities across diverse complexity scenarios.

To assess the models' ability to solve problems with multiple attempts, we further evaluate their performance under Pass@3, as shown in Table~\ref{tab:pass@3}.  The results show that Pass@3 consistently outperforms Pass@1, demonstrating enhanced problem-solving capabilities when models are given more opportunities. Despite the overall performance gain, the relative ranking of models remains largely unchanged, confirming the stability of our benchmark. Additionally, the varying degrees of improvement across complexity levels reveal the models’ differing abilities to handle increasingly complex tasks.

\begin{table*}[th!]
\definecolor{deepgreen}{RGB}{0,100,0}
  \centering
  
    \small
  \renewcommand{\arraystretch}{0.95}
  \resizebox{\linewidth}{!}{ 
  \begin{tabular}{lr|cc|ccccc}
    \toprule
    
    \multirow{2}{*}{\textbf{Model}} & \multirow{2}{*}{\textbf{Params}}  & \multirow{2}{*}{\textbf{MBPP}} & \multirow{2}{*}{\textbf{MBPP+}}   & \multicolumn{5}{c}{\textbf{DynaCode}} \\
    & & & & \textbf{Unit 1} & \textbf{Unit 2} & \textbf{Unit 3} & \textbf{Unit 4} & \textbf{Average} \\
    \midrule

    GPT-4o & - & 91.0 & 78.6 & 79.9 & 62.9 & 65.4 & 51.9 & 65.0 \\
    GPT-3.5-Turbo & - & 86.2 & 75.1 & 42.4 & 39.3 & 34.1 & 35.6 & 37.8 \\
    \midrule 
    
    DeepSeek-V3 & 236B & 88.9 & 74.6 & 72.5 & 49.4 & 65.8 & 57.8 & 61.4 \\
    \midrule
    
    Qwen2.5-Coder-32B-Instruct & 32B & 92.9 & 81.5 & 68.8 & 44.9 & 56.6 & 44.3 & 53.7 \\
    \midrule
    
    WizardLM-2-8x22B & 176B & 77.0 & 67.5 & 42.9 & 36.7 & 33.4 & 17.4 & 32.6 \\
    \midrule
    
    Mixtral-8x22B-Instruct-v0.1 & 176B & 78.0 & 66.4 & 43.3 & 37.9 & 33.3 & 17.8 & 33.1 \\
    \midrule
    
    Phind-CodeLlama-34B-v2 & 34B & 89.9 & 76.5 & 46.1 & 36.9 & 58.3 & 32.9 & 43.6 \\
    \midrule
    
    starcoder2-15b-instruct-v0.1 & 15B & 90.2 & 77.2 & 46.1 & 36.4 & 59.1 & 31.6 & 43.3 \\
    \midrule
    
    codegemma-7b-it & 7B & 89.2 & 77.2 & 8.1 & 5.3 & 2.9 & 6.3 & 5.6 \\
    \midrule
    
    Meta-Llama-3.1-405B-Instruct & 405B & 91.0 & 77.5 & 57.3 & 47.5 & 56.9 & 34.8 & 49.1 \\
    Meta-Llama-3.3-70B-Instruct & 70B & 91.5 & 78.6 & 42.3 & 34.3 & 67.1 & 38.9 & 45.6 \\
    Meta-Llama-3.1-8B-Instruct & 8B & 82.8 & 73.0 & 21.9 & 14.8 & 15.6 & 14.2 & 16.6 \\

    \bottomrule
  \end{tabular}
  }
  \caption{Pass@3 results on MBPP, MBPP+, and our DynaCode across varying code complexities. }
  \label{tab:pass@3}
\end{table*}

\subsection{Details of the Number}
\label{app:performance_number}

\begin{figure*}[t]
\centering
\includegraphics[width=1\textwidth]{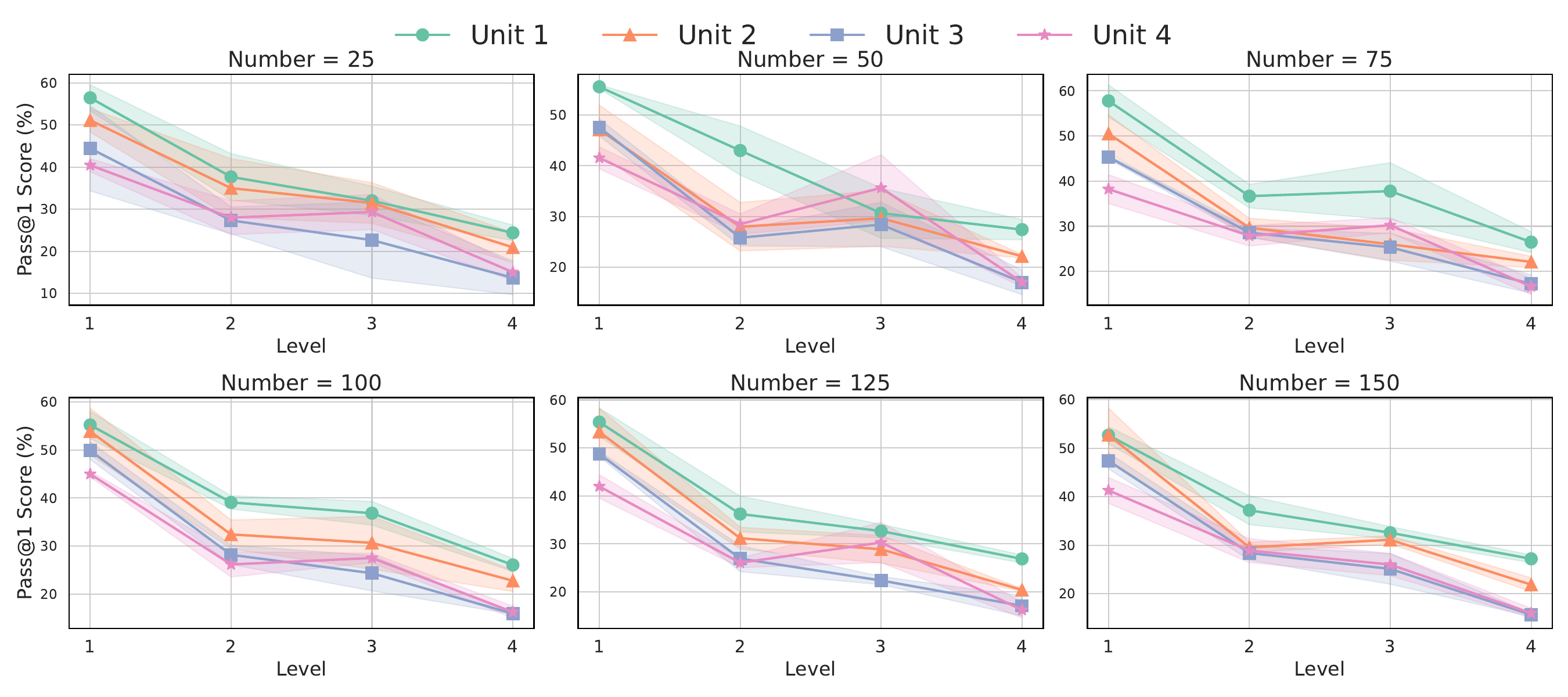}
\caption{Performance comparison of GPT-3.5-Turbo on DynaCode with varying numbers of dynamically generated problems $\{25, 50, 75, 100, 125, 150\}$.}
\label{fig:Number_datails}
\end{figure*}

In Figure~\ref{fig:Number_datails}, we present the detailed results of GPT-3.5's performance for each unit, with problem quantities set to $\{25, 50, 75, 100, 125, 150\}$ for each graph.
The results indicate that smaller $N$ values, such as 25 and 50, lead to greater performance variability across units, suggesting sensitivity to insufficient data. As $N$ increases, especially beyond 75, the performance stabilizes, and fluctuations diminish. Based on this observation, $100$ was selected for subsequent experiments, balancing computational efficiency with evaluation robustness.

\begin{table}[t!]
\small
\centering
\resizebox{\linewidth}{!}{ 
\begin{tabular}{l|cccc}
\toprule
\textbf{Type} & \textbf{Unit 1} & \textbf{Unit 2} & \textbf{Unit 3} & \textbf{Unit 4} \\
\midrule
Base     & 153     & 100     & 76 & 76     \\
\midrule
Graph 1    & 2,617          & 697  & 568          & 1,026    \\
Graph 2    & 48,638         & 5,718 & 4,133         & 13,517   \\
Graph 3    & 37,084         & 4,138 & 3,230         & 9,453    \\
Graph 4    & 896,634        & 43,051 & 26,648       & 166,466  \\
Graph 5    & 710,792        & 34,101 & 19,768       & 120,555  \\
Graph 6    & 1,342,944       & 64,915 & 48,304       & 242,107  \\
Graph 7    & 349,802        & 15,883 & 12,858       & 56,251   \\
Graph 8    & 15,938,962      & 288,317 & 150,618     & 1,931,587 \\
Graph 9    & 12,796,855      & 241,443 & 108,883     & 1,423,255 \\
Graph 10   & 37,303,818      & 695,172 & 601,704     & 4,189,592 \\
Graph 11   & 24,051,797      & 458,375 & 320,522     & 2,881,061 \\
Graph 12   & 11,832,328      & 229,898 & 181,992     & 1,497,380 \\
Graph 13   & 19,069,749      & 362,085 & 237,205     & 2,083,414 \\
Graph 14   & 710,792        & 34,101  & 19,768      & 120,555  \\
Graph 15   & 2,426,580       & 42,459  & 39,678      & 239,392  \\
Graph 16   & 36,998,540      & 695,172 & 600,528     & 4,177,666 \\
\midrule
Unit Sum  & 164,517,932 & 3,215,525  & 2,376,407 & 19,153,277  \\
Total   &  \multicolumn{4}{c}{189,263,141}     \\
\bottomrule
\end{tabular}
}
\caption{The number of problems contained in different units and the corresponding number of benchmarks that can be generated in combination.}
\label{tab:graph_number_details}
\end{table}

\begin{table}[t!]
\small
\centering
\begin{tabular}{l|c}
\toprule
\textbf{Benchmark} & \textbf{Number of Problems} \\
\midrule
HumanEval     & 164       \\
HumanEval+  & 164   \\
MBPP  & 974    \\
MBPP+  & 378    \\
BigCodeBench & 1,140  \\
\midrule
DynaCode & 189,263,141     \\
\bottomrule
\end{tabular}
\caption{Comparison of the number of problems between DynaCode and other code generation benchmarks.}
\label{tab:benchamrk_size}
\end{table}

\section{Prompt}
\label{app:prompt}

In our DynaCode, we dynamically generate a unified prompt by hard-coding the combination of individual code problems. Initially, a call graph is constructed by selecting candidate functions based on their input and output types. The graph is then completed by traversing it and randomly assigning nodes while ensuring that the data flow remains consistent. Specifically, the output of a parent function becomes the input for its child function. Each node is assigned a unique prompt number, and the individual prompts stored in our dataset are concatenated in sequence to form the final prompt. 

Table~\ref{tab:prompt_example} presents an example from $G_8$, illustrating this process in practice. In addition to the sequential prompt assembly, a main function is automatically generated to call the individual functions in the correct order, with explicit rules that govern the data transfer between them. This hard-coded dynamic composition strategy not only ensures logical consistency among the code fragments but also enhances the robustness and adaptability of our code generation system.

\section{Error Classification Details}
\label{app:error}
We present the classification of error types in DynaCode along with their corresponding functional details, as shown in Table~\ref{tab:error_categorization}.

\begin{table*}[th!]
\centering
\small
\renewcommand{\arraystretch}{1.1}
\resizebox{\linewidth}{!}{
\begin{tabular}{l|l|l}
\toprule
\textbf{Associated Capability} & \textbf{Error Type} & \textbf{Reason for Categorization} \\
\midrule
Problem Understanding & AssertionError & Failure to satisfy the expected logic of the problem, indicating misunderstanding of requirements. \\
Problem Understanding & ValueError & Input data out of expected range, implying insufficient understanding of problem constraints. \\
Problem Understanding & RecursionError & Incorrect handling of recursion logic, showing failure in comprehending recursive termination. \\
Problem Understanding & ZeroDivisionError & Failure to account for boundary conditions, such as division by zero. \\
\midrule
Code Pattern Generation & SyntaxError & Violation of syntax rules, directly reflecting the inability to generate structurally correct code. \\
Code Pattern Generation & IndentationError & Incorrect indentation, indicating issues in formatting and structural management of code. \\
\midrule
Context Management & NameError & Use of undefined variables, highlighting issues in variable scope management. \\
Context Management & AttributeError & Accessing nonexistent attributes, suggesting misunderstanding of object structures. \\
Context Management & TypeError & Incorrect data type handling, indicating flaws in type inference and function parameter management. \\
Context Management & IndexError & Indexing beyond valid range, showing poor management of data structures like lists or arrays. \\
Context Management & UnboundLocalError & Use of uninitialized local variables, indicating incorrect variable lifecycle management. \\
\midrule
Other & OverflowError & Numeric overflow due to improper handling of large values or operations. \\
Other & RuntimeError & Errors during execution, often caused by incorrect function calls or resource issues. \\

\bottomrule
\end{tabular}
}
\caption{Categorization of error types and their corresponding capabilities in DynaCode.}
\label{tab:error_categorization}
\end{table*}

\section{Examples in DynaCode}
\label{app:example}
\begin{figure*}[t]
\centering
\includegraphics[width=0.85\textwidth]{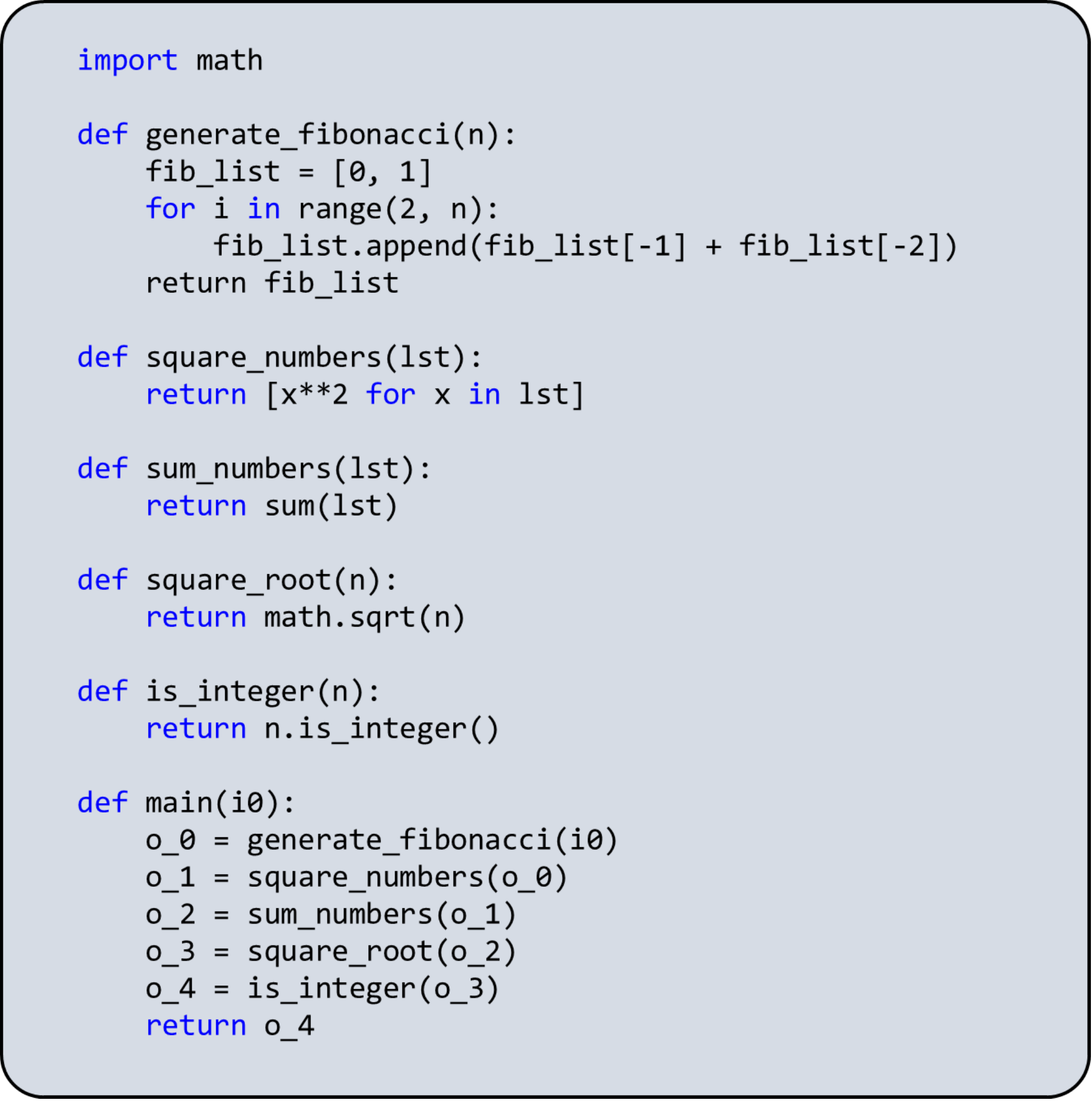}
\caption{
Example of a nested code workflow from Table~\ref{tab:prompt_example}, demonstrating how the call graph-based evaluation assesses LLMs' ability to handle multi-step tasks and dependencies.
}
\label{fig:code_example}
\end{figure*}

We provide a valid nested code example for the prompt in Table~\ref{tab:prompt_example}, as shown in Figure~\ref{fig:code_example}. As observed, the flow from \textit{generate\_fibonacci} to \textit{is\_integer} represents a complete workflow, which demonstrates that the call graph-based dynamic evaluation strategy can assess the ability of LLMs to handle multi-step tasks and the dependencies between different steps by simulating a real-world task flow.
Our call-graph structure offers a variety of workflow configurations, which can help us understand how well the LLM performs on tasks requiring structured thinking and logical progression, making it a powerful tool for evaluating its capabilities in multi-step tasks.

\begin{figure*}[t]
\centering
\includegraphics[width=0.9\textwidth]{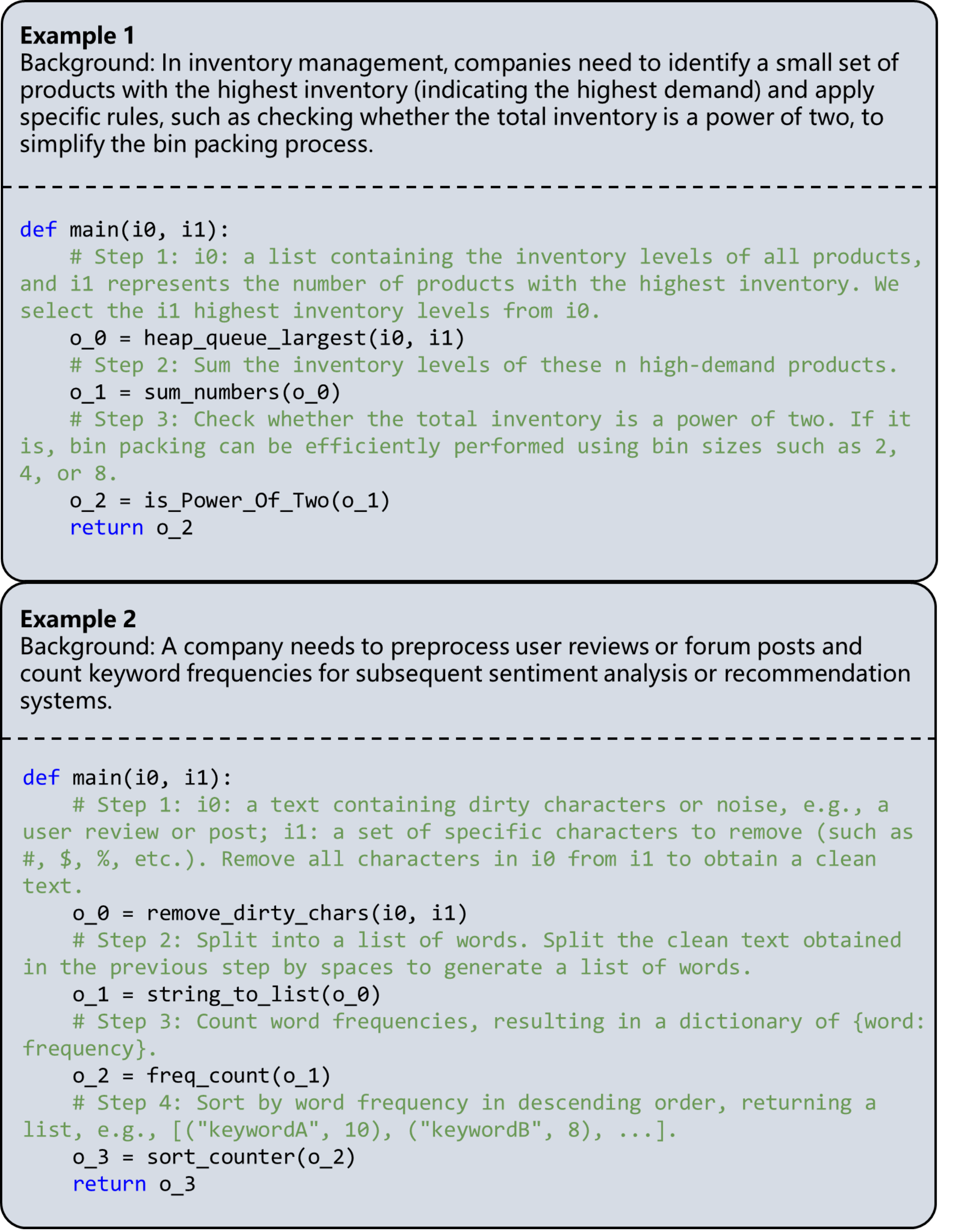}
\caption{
Example 1 (call graph $G_2$) and Example 2 (call graph $G_4$) illustrate realistic multi-step tasks in inventory analysis and text preprocessing. This demonstrates how our call-graph structures capture interdependent sub-tasks commonly found in real-world coding scenarios. }
\label{fig:examples_in_reality}
\end{figure*}

To further demonstrate the practicality of our benchmark, we present two representative examples in Figure~\ref{fig:examples_in_reality}. These cases are grounded in realistic coding scenarios, such as text preprocessing and inventory analysis, and illustrate how our call-graph structure captures diverse, interdependent sub-tasks commonly encountered in real-world applications.

\begin{table*}[!t]
    \centering
    
    \setlength\tabcolsep{3pt}
    \resizebox{1.0\textwidth}{!}{%
    \begin{tcolorbox}[colback=blue!5!white, colframe=black, width=1.0\textwidth, title={Python Code Prompts}]
\small
Here are 5 prompts that are used to generate 5 functions respectively.\\[1ex]

\textbf{PROMPT 1:}\\
\texttt{""" }\\
Write a python function to generate the first n Fibonacci numbers.\\
\texttt{assert generate\_fibonacci(5) == [0, 1, 1, 2, 3]}\\
\texttt{""" }\\[1ex]

\textbf{PROMPT 2:}\\
\texttt{""" }\\
Write a python function to square each number in a given list.\\
\texttt{assert square\_numbers([0, 1, 1, 2, 3]) == [0, 1, 1, 4, 9]}\\
\texttt{""" }\\[1ex]

\textbf{PROMPT 3:}\\
\texttt{""" }\\
Write a python function to find the sum of all numbers in a list.\\
\texttt{assert sum\_numbers([0, 1, 1, 4, 9]) == 15}\\
\texttt{""" }\\[1ex]

\textbf{PROMPT 4:}\\
\texttt{""" }\\
Write a python function to find the square root of a number.\\
\texttt{assert square\_root(15) == 3.872983346207417}\\
\texttt{""" }\\[1ex]

\textbf{PROMPT 5:}\\
\texttt{""" }\\
Write a python function to check if a number is an integer.\\
\texttt{assert is\_integer(3.872983346207417) == False}\\
\texttt{""" }\\[1ex]

Please write the above 5 functions respectively and write a new function named \texttt{main} to call the above 5 functions.\\[1ex]

When calling these functions, please follow the following rules:\\

The input of the main function equals the input of PROMPT 1 :\texttt{generate\_fibonacci}.\\

The output of function PROMPT 1: \texttt{generate\_fibonacci} serves as the input of PROMPT 2: \texttt{square\_numbers}.\\

The output of function PROMPT 2: \texttt{square\_numbers} serves as the input of PROMPT 3: \texttt{sum\_numbers}.\\

The output of function PROMPT 3: \texttt{sum\_numbers} serves as the input of PROMPT 4: \texttt{square\_root}.\\

The output of function PROMPT 4: \texttt{square\_root} serves as the input of PROMPT 5: \texttt{is\_integer}.\\[1ex]

The main function returns the output of the PROMPT 5: \texttt{is\_integer}.

    \end{tcolorbox}
    }
    \caption{Example of a dynamically generated prompt in DynaCode for call graph $G_8$.}
    \label{tab:prompt_example}
\end{table*}

\end{document}